\documentclass{amia}
\usepackage{graphicx}
\usepackage[export]{adjustbox}
\usepackage[labelfont=bf]{caption}
\usepackage[superscript,nomove]{cite}
\usepackage[dvipsnames]{xcolor}
\usepackage{multirow}
\usepackage{url}
\usepackage{soul}
\usepackage[symbol]{footmisc}
\usepackage[square,sort,numbers,super]{natbib}
\usepackage{cleveref}

\crefformat{section}{\S#2#1#3}
\crefformat{subsection}{\S#2#1#3}
\crefformat{subsubsection}{\S#2#1#3}

\newcommand\EatDot[1]{}

\begin{document}

\title{Extracting Concepts for Precision Oncology from the Biomedical Literature}

\author{Nicholas Greenspan,$^1$\footnote[1]{This project was undertaken during an undergraduate internship at UTHealth-SBMI.} Yuqi Si, MS,$^2$ Kirk Roberts, PhD$^2$}

\institutes{
    $^1$Department of Computer Science, Rice University\\
    Houston TX, USA \\
    $^2$School of Biomedical Informatics,\\
    The University of Texas Health Science Center at Houston \\
    Houston TX, USA \\
}

\maketitle

\noindent{\bf Abstract}

\textit{%
This paper describes an initial dataset and automatic natural language processing (NLP) method for extracting concepts related to precision oncology from biomedical research articles.
We extract five concept types: {\sc Cancer}, {\sc Mutation}, {\sc Population}, {\sc Treatment}, {\sc Outcome}.
A corpus of 250 biomedical abstracts were annotated with these concepts following standard double-annotation procedures.
We then experiment with BERT-based models for concept extraction.
The best-performing model achieved a precision of 63.8\%, a recall of 71.9\%, and an F1 of 67.1.
Finally, we propose additional directions for research for improving extraction performance and utilizing the NLP system in downstream precision oncology applications.
}

\section{Introduction}
\label{section:introduction}

Precision medicine is a paradigm in which treatment decisions are based not just on a patient's disease status, but on a variety of other factors including specific genetic, environmental, and other factors.\cite{Collins15}
The preeminent use case for precision medicine thus far has been cancer, i.e. precision oncology.
Precision oncology is a rapidly-developing field\cite{Garraway18}, with a growing number of treatments, trials, and genomic markers.
Since drugs can be targeted to relatively rare mutations, the number of studied treatments is greatly expanded \cite{Hewett02,Barbarino18} and these can be referred to by a variety of names (e.g., the name used in pre-clinical trials is often different than the final drug name).
Since the gene mutations can be relatively rare, clinical trial structures have had to be altered to better fit the precision medicine paradigm.\cite{Fountzilas18}
And, critically, there are thousands of known genetic mutations from hundreds of cancer-related genes.\cite{Chakravarty17}
Sizable effort is thus required to curate all of these types of information to make them available in a usable form to both researchers and clinicians.

Our prior work has focused on this problem from an information retrieval (IR) perspective: how does one find patient-specific information (given a type of cancer, mutation, etc.) from the vast trove of precision medicine-related publications.
IR systems were evaluated for this task in the TREC Precision Medicine tracks.\cite{Roberts17.trec,Roberts18.trec,Roberts19.trec}
We also developed PRIMROSE \cite{Shenoi20.amia_summit}, a search engine that implements many of the best aspects of precision oncology search.
A consistent weakness in these IR approaches, however, was difficulty dealing with the complex semantics of precision oncology articles: identifying the exact treatments studied in an article, which types of cancer the treatment applies to, etc.
This task is more consistent with a natural language processing (NLP) information extraction (IE) approach.
Therefore, in this work we report the initial development of an NLP system for extracting five key elements of biomedical articles for precision oncology: the type(s) of cancer studied, the mutations that were targeted, the specific population it is limited to, the treatment evaluated, and any available outcome information summarized in the abstract.
Because of the fast-moving nature of the field, we focus on biomedical abstracts instead of full-text articles.
Not only are the abstracts publicly available well before the full text, but many of the latest-breaking developments in precision oncology are presented at talks in major oncology conferences and only the abstracts for these talks are provided.

To gauge the complexity of this NLP task, we collected a pilot corpus of 250 biomedical abstracts drawn from the TREC Precision Medicine dataset.
The five concept types--{\sc Cancer}, {\sc Mutation}, {\sc Population}, {\sc Treatment}, and {\sc Outcome}--were double-annotated and reconciled.
Two models based on BERT\cite{Devlin18}, and specifically the BioBERT\cite{Lee20} model pre-trained on biomedical text, were evaluated: BioBERT$_{\textrm{{\sc base}}}$ and BioBERT$_{\textrm{{\sc large}}}$.
The difference between these models is the number of parameters, in terms of number of layers, hidden units, and attention heads.

The remainder of this paper is organized as follows.
Section~\ref{section:background} discusses related work in NLP for cancer and precision medicine.
Section~\ref{section:methods} describes the methods, including data (\cref{subsection:data}), annotation (\cref{subsection:annotation_process}), and automatic concept extraction (\cref{subsection:automatic_extraction}).
Section~\ref{section:results} details the results.
Section~\ref{section:discussion} provides a discussion, including an error analysis, implications, and directions for future work.
Finally, Section~\ref{section:conclusion} concludes the paper.

\section{Related Work}
\label{section:background}

\paragraph{Biomedical Literature NLP for Cancer}

Cancer is one of the more frequently studied aspects of NLP for biomedical literature articles.
Early works such as MedScan\cite{Novichkova03} employed rule-based systems to extract and interpret information from MEDLINE abstracts.
Chun et al.\cite{Chun06} developed a corpus and extracted relations between prostate cancer and genes from abstracts using a maximum entropy classifier.
Baker et al. developed a corpus for identifying the hallmarks of cancer from the biomedical literature and proposed a support vector machine (SVM) model\cite{Baker16.bioinformatics} and later a convolutional neural network\cite{Baker16.biotxtm} to automatically classify abstracts.
A different take on cancer NLP for the biomedical literature is the development of literature-based discovery (LBD) tools such as LION LBD\cite{Pyysalo19.bioinformatics} to identify implicit links within the network of literature articles.
LION in particular focuses on the molecular biology of cancer.
Beyond the biomedical literature, a tremendous amount of NLP research has been conducted for cancer on other data types.
Most notable among these are electronic health records, for which several review articles exist that overview cancer NLP for clinical notes.\cite{Spasic14,Yim16,Datta19.jbi} 

\vspace{-0.1in}

\paragraph{Biomedical Literature NLP for Genomics}
A tremendous amount of NLP work has focused on extracting information related to genomics from the literature.
Early work includes EDGAR\cite{Rindflesch00.biocomputing} 
, which identified gene-drug relations from biomedical abstracts.
Libbus et al.\cite{Libbus04} identified genes from MEDLINE abstracts based on the Gene Ontology\cite{GOConsortium04} for the purpose of linking literature-based data to structured knowledge sources.
Work in pharmacogenomics has required extensive use of NLP to build resources such as the use of SemRep\cite{Ahlers07} or the construction of the pharmacogenomics knowledge base PharmGKB. \cite{Klein01,WhirlCarrillo12,Thorn13}
In turn, PharmGKB has been utilized as a knowledge base for many further NLP studies.\cite{Pakhomov12,Buyko12,Ravikumar14} 
Similarly, the PGxCorpus\cite{Legrand20} is a manually-annotated corpus for pharmacogenomics--similar in many ways to our goal here, but their work is not specific to cancer.
Finally, more general biomedical literature NLP has included genomic components, particularly the
CRAFT corpus.\cite{Cohen17.hla,Baumgartner19}

\vspace{-0.1in}

\paragraph{Biomedical Literature NLP for Precision Oncology}

There has indeed been some work specific to precision medicine for NLP within the space of the current work.
For instance, Deng et al.\cite{Deng19.jcocci} classifies abstracts with an SVM based on whether they focus on cancer penetrance.
Bao et al.\cite{Bao19.jcocci} extends this with a deep learning model. 
Instead of extracting the particular concepts, however, these works focus is simply to classify the entire abstract for use in downstream meta-analyses.
Next, Hughes et al.\cite{Hughes20.breast} reviews how to utilize precision oncology NLP specific for breast cancer.
Finally, the TREC Precision Medicine track\cite{Roberts17.trec,Roberts18.trec,Roberts19.trec} is an ongoing information retrieval shared task focusing on identifying articles relevant to precision oncology.
This has inspired the creation of many search engines, including our own,\cite{Shenoi20.amia_summit} for clinical decision support in precision oncology.
Of the many search engines to participate in the TREC Precision Medicine track, however, none has successfully integrated biomedical knowledge sources to greatly improve retrieval performance.
We believe this is partly due to the fact that it is difficult to properly link the key aspects of precision oncology in an abstract to these powerful knowledge bases.
Instead, most use of biomedical knowledge in such search engines is simply to expand synonyms (e.g., through query expansion) which gives at most small boosts to retrieval performance.
Our goal in this paper, then, is to lay the groundwork for improvements in precision oncology search and knowledge acquisition by identifying the key elements to precision oncology in biomedical abstracts.
This will allow for the downstream linking of these articles with existing biomedical knowledge bases for better semantic comprehension of the precision oncology scientific landscape.

\section{Methods}
\label{section:methods}

The high-level study design for this paper follows the standard supervised NLP pipeline: data identification (Section~\ref{subsection:data}), manual data annotation (Section~\ref{subsection:annotation_process}), and automatic NLP extraction (Section~\ref{subsection:automatic_extraction}).
Since this is a pilot study, our primary goal has been to identify the key barriers to large-scale system development, which is discussed in more detail in the Discussion (Section~\ref{section:discussion}).

\subsection{Data}
\label{subsection:data}

Since the latest developments of precision oncology research are only publicly available in abstracts, we focus only on abstract-based annotation and extraction.
Compared to biomedical research in general, precision oncology is disproportionately less represented in PubMed Central given its funding structure (less open access, more embargoed journal articles) and heavy use of abstract presentations for presenting results--which means many of the latest developments that are so important to capture are not available as full text articles, but only abstracts.
We focus on a set of abstracts known to be relevant to precision oncology by annotating only abstracts judged as relevant in the TREC 2017 Precision Medicine track\cite{Roberts17.trec}.
A random selection of 250 abstracts was chosen from those judged relevant during the assessment process.

\subsection{Annotation Process}
\label{subsection:annotation_process}

The 250 abstracts were imported into Brat\cite{Stenetorp12} and double-annotated with the following concept types:

\begin{enumerate}
    \item {\sc Cancer}. The type of cancer being studied in the article (e.g., ``{\em breast cancer}'', ``{\em non-small cell lung cancer}'', ``{\em mantle cell lymphoma}'', ``{\em solid tumor}''). If the abstract mentions a type of cancer but it is clearly not the cancer investigated in the study, then it is additionally labeled as a Non-study cancer. If multiple types of cancer are included in the study, all are annotated.
    
    \item {\sc Mutation}. The gene mutation being studied in the article, be it a gene with any mutation (e.g., ``{\em KRAS}'', ``{\em FGFR2}'', ``{\em PIK3R1}''), a specific variant (e.g., ``{\em BRAF V600E}'', ``{\em KRAS G13D}'', ``{\em NF2 K322}''), or some other form of genetic mutation (e.g., ``{\em CDK4 Amplification}'', ``{\em PTEN Inactivating}'', ``{\em EML4-ALK Fusion transcript}'').
    Similar to cancer type, mutations mentioned in the abstract but not investigated in the study are marked as Non-study mutations.
    
    \item {\sc Population}. The specific population in the study (e.g., ``{\em Hunan Province in China}'', ``{\em never or light smokers}'', ``{\em adults ($>$ 18 years)}'', ``{\em European patients}'', ``{\em no history of chemotherapy for metastatic disease}'').
    As shown by the examples, this can include age, sex, location, ethnicity, cancer status, etc.
    Populations mentioned in the abstract but not investigated in the study are marked as Non-study populations.
    
    \item {\sc Treatment}. The drug used in the study (e.g., ``{\em sorafenib}'', ``{\em abemaciclib}'', ``{\em trastuzumab}'').
    If the drug was used as part of a combination, each individual component is annotated separately.
    If the drug was a comparator but not directly investigated in the study, then it is marked as a Non-study treatment (this is more common than Non-study cancers, mutations, and populations).
    
    \item {\sc Outcome}. The result of the study with regards to the success or failure of the treatment.
    Non-study outcomes are not annotated.
    The outcomes are generally a sentence or long phrase describing the overall outcome.
    E.g.,
    \begin{itemize}
        \item {\em Main grade 3 or 4 toxicities were rash (11 [13\%] of 84 patients given erlotinib vs none of 82 patients in the chemotherapy group), neutropenia (none vs 18 [22\%]), anaemia (one [1\%] vs three [4\%]), and increased amino-transferase concentrations (two [2\%] vs 0).}
        \item {\em Treatment with crizotinib results in clinical benefit rate of 85\%-90\% and a median progression-free survival of 9-10 months for this molecular subset of patients.}
        \item {\em Although nearly all patients with GIST treated with imatinib experienced adverse events, most events were mild or moderate in nature.}
    \end{itemize}
\end{enumerate}

Additionally, negated concepts were marked as such.
While there were negated {\sc Cancer} annotations (e.g., 

Two annotators (the first author and a biomedically-trained graduate student) labeled each abstract in batches of 25, reconciling after each batch.
Instead of using highly-refined guidelines, the goal of this annotation process was more exploratory in nature.
The concepts were defined as above, but no further.
The goal was to identify the range of possible ways in which the information can be expressed, without too much regard for maximizing inter-rater agreement.
Anecdotally, some concepts had more inconsistent agreement throughout the process (notably {\sc Population} and {\sc Outcome}), while others had early disagreement that improved over time (such as how to handle acronyms with {\sc Cancer} and {\sc Mutation}).
These issues are ultimately reflected in the automatic extraction scores described in Section~\ref{section:results}.

\begin{figure}[p]
\centering
\includegraphics[width=\linewidth,frame]{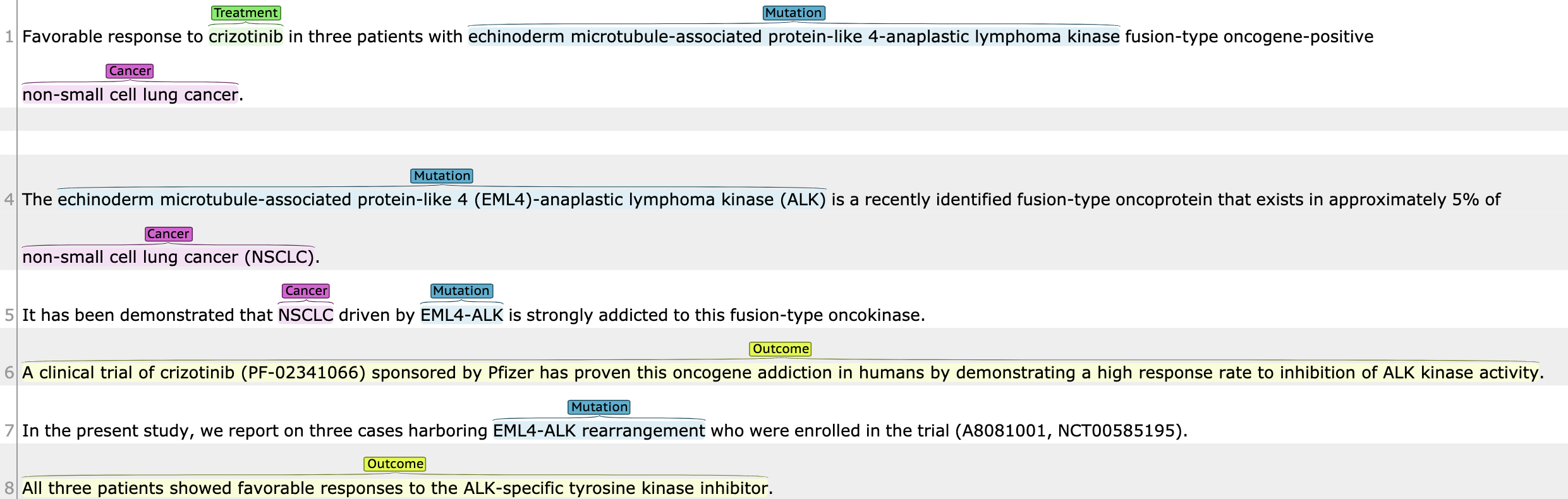}\\
\vspace{0.1in}
\includegraphics[width=\linewidth,frame]{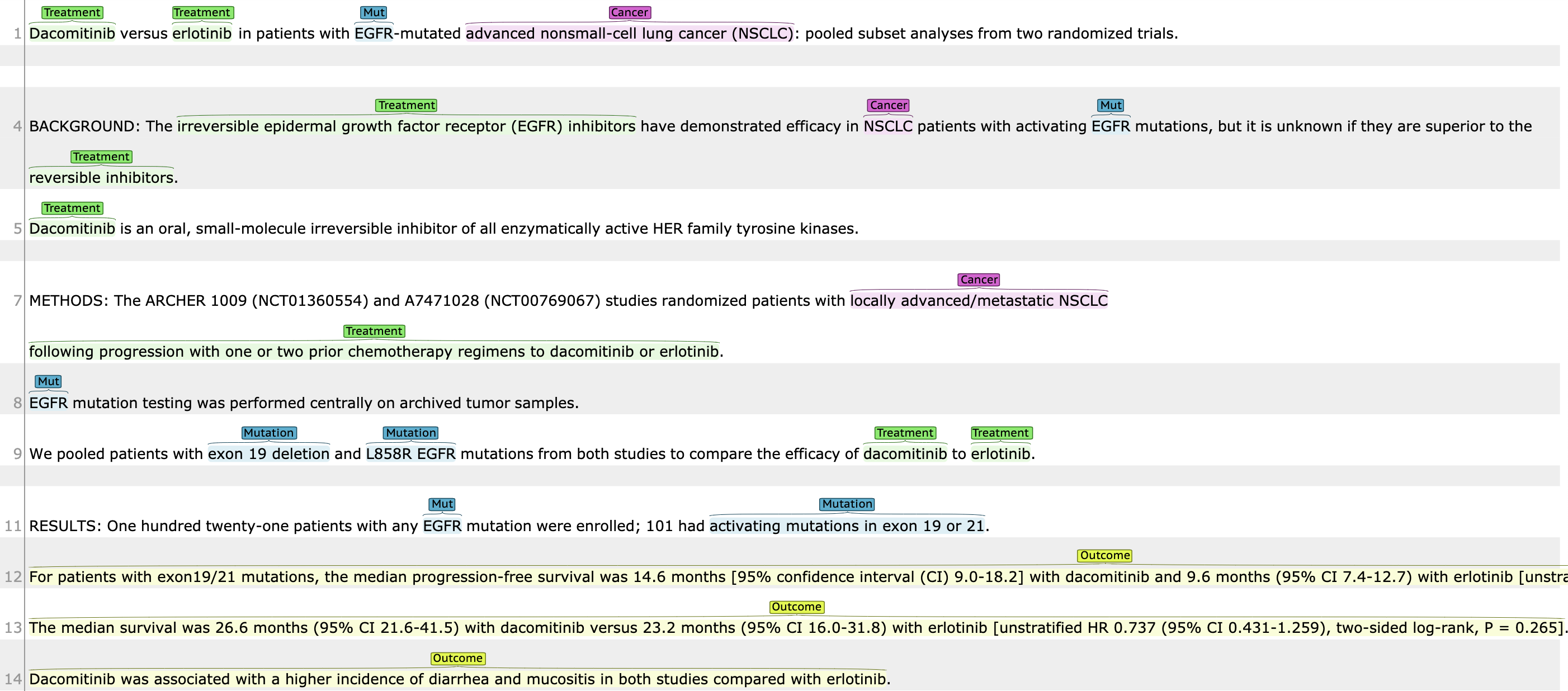}\\
\vspace{0.1in}
\includegraphics[width=\linewidth,frame]{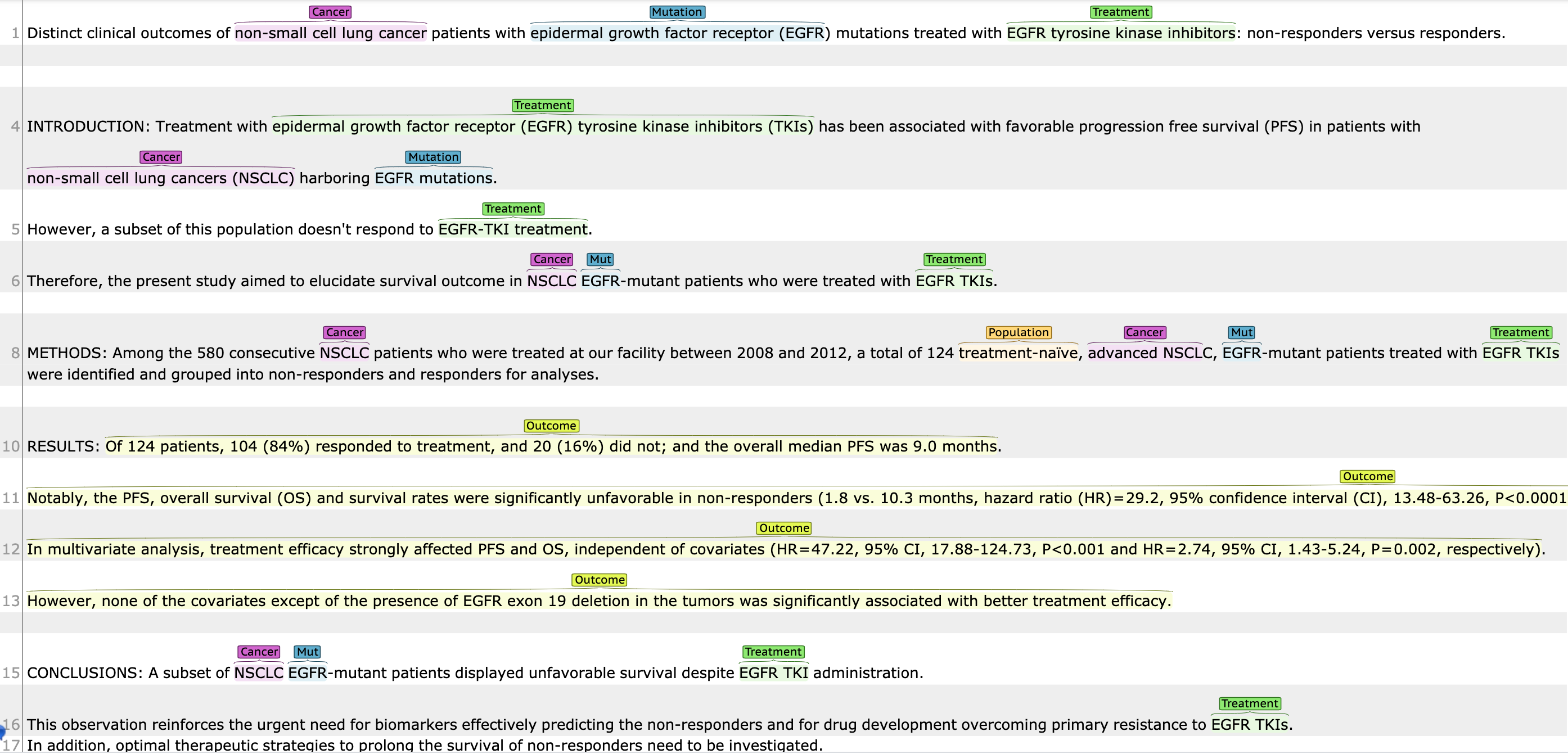}
\caption{Example annotations}
\label{figure:examples}
\end{figure}

Descriptive statistics of the annotated corpus are provided in Table~\ref{table:descriptive_stats}.
Example annotations from the corpus are shown in Figure~\ref{figure:examples}.

\begin{table}[t]
\centering
\begin{tabular}{ll}
\hline
Number of abstracts & 250 \\
Average length of abstract (tokens) & 278.1 \\
\hline
Total concept annotations & 4,722 \\
~~~~{\sc Cancer}  & 1,622 \\
~~~~{\sc Mutation}  & 2,293 \\
~~~~{\sc Population}  & 133 \\
~~~~{\sc Treatment}  & 544 \\
~~~~{\sc Outcome}  & 130 \\
\hline
Percent Non-study annotations & 1.2\% \\
~~~~{\sc Cancer}  & 0.9\% \\
~~~~{\sc Mutation}  & 0.8\% \\
~~~~{\sc Population}  & 0.8\% \\
~~~~{\sc Treatment}  & 4.0\% \\
~~~~{\sc Outcome}  & 0.0\% \\
\hline
Average concept length (tokens) & 3.3 \\
~~~~{\sc Cancer}  & 2.7 \\
~~~~{\sc Mutation}  & 2.3 \\
~~~~{\sc Population}  & 4.4 \\
~~~~{\sc Treatment}  & 3.0 \\
~~~~{\sc Outcome}  & 28.5 \\
\hline
\end{tabular}
\caption{Descriptive statistics of the annotated corpus.}
\label{table:descriptive_stats}
\end{table}

\subsection{Automatic Extraction}
\label{subsection:automatic_extraction}

The abstracts were tokenized and split into sentences using spaCy \cite{spacy2}.
A BILOU scheme was used for sequence classification, where B is the first token of a sequence, I an inside token, L the last token, O a token outside any sequence, and U a single-token concept.
So ``{\em K - ras and PTEN mutations}'' would be [{\sc B-Mutation}, {\sc I-Mutation}, {\sc L-Mutation}, {\sc O}, {\sc U-Mutation}, {\sc O}].
Non-study concepts were handled by adding a {\sc N-} before the concept name (e.g., {\sc B-N-Treatment}).

We follow the standard BERT framework for named entity recognition tasks.
Two variants of BioBERT\cite{Lee20} were evaluated: BioBERT$_{\textrm{{\sc base}}}$ v1.1 and BioBERT$_{\textrm{{\sc large}}}$ v1.1,
which are versions of BERT$_{\textrm{{\sc base}}}$ and BERT$_{\textrm{{\sc large}}}$ respectively pre-trained on both 1 million PubMed abstracts (note that the BioBERT v1.0 models are pre-trained on 200k PubMed abstracts and 200k PubMed Central full-text articles, but BioBERT v1.1 is only pre-trained on abstracts, though a larger number).
As such, BioBERT is an ideal starting point for a transformer-based language model to use for our task.
BioBERT$_{\textrm{{\sc base}}}$ has 12 layers, 768 hidden units per layer, and 12 attention heads per layer (a total of 110 million parameters);
BioBERT$_{\textrm{{\sc large}}}$ has 24 layers, 1024 hidden units per layer, and 16 attention heads per layer (a total of 340 million parameters).
Generally, the larger BERT variant offers some improved performance, but in many cases the performance delta is neglible and not worth the additional computational cost.
As such, we experiment with both models to assess whether a larger BERT model would be beneficial in this task.

The data was split 70\% for training the BioBERT models, 10\% for validation (early stopping), and 20\% for testing (results discussed below).
The default BioBERT parameters were used other than a learning rate of $2x10^{-5}$, maximum sequence length of 128, training batch size of 32, validation batch size of 8, and test batch size of 8.

\newpage

\section{Results}
\label{section:results}

\begin{table}[t]
\centering
\vspace{0.3in}
\begin{tabular}{l|ccc}
Annotation & Precision & Recall & F1 \\
\hline
Overall          & 60.48 & 70.73 & 65.20 \\
\hline
{\sc Cancer}     & 69.31 & 78.65 & 73.68 \\
{\sc Mutation}   & 59.35 & 69.13 & 63.87 \\
{\sc Population} & 41.82 & 42.59 & 42.20 \\
{\sc Treatment}  & 47.79 & 71.05 & 57.14 \\
{\sc Outcome}    & 0.0   & 0.0   & 0.0 \\
\hline
\end{tabular}
\caption{Results using BioBERT$_{\textrm{{\sc base}}}$ model.}
\label{table:base_results}
\end{table}

\begin{table}[t]
\centering
\vspace{0.3in}
\begin{tabular}{l|ccc}
Annotation & Precision & Recall & F1 \\
\hline
Overall          & 63.79 & 71.90 & 67.61 \\
\hline
{\sc Cancer}     & 70.54 & 80.06 & 75.00 \\
{\sc Mutation}   & 61.51 & 68.78 & 64.94 \\
{\sc Population} & 56.25 & 50.00 & 52.94 \\
{\sc Treatment}  & 58.59 & 76.32 & 66.29 \\
{\sc Outcome}    & 0.0   & 0.0   & 0.0 \\
\hline
\end{tabular}
\caption{Results using BioBERT$_{\textrm{{\sc large}}}$ model.}
\label{table:large_results}
\end{table}

The results for the BioBERT$_{\textrm{{\sc base}}}$ and BioBERT$_{\textrm{{\sc large}}}$ models are provided in Table~\ref{table:base_results} and Table~\ref{table:large_results}.
Not enough Non-study concepts are present in the test set to merit an evaluation here.
We thus focus on boundary extraction and type classification without the Non-study attribute.

In almost every case, the BioBERT$_{\textrm{{\sc large}}}$ results outperform the BioBERT$_{\textrm{{\sc base}}}$ results (the lone exception being {\sc Mutation} recall, while neither model successfully extracts any {\sc Outcome}).
The differences between BioBERT$_{\textrm{{\sc base}}}$ and BioBERT$_{\textrm{{\sc large}}}$ are often several points, including substantial boosts for both {\sc Population} (+10.74 F1) and {\sc Treatment} (+9.15 F1).
Notably, the improvements from BioBERT$_{\textrm{{\sc base}}}$ to BioBERT$_{\textrm{{\sc large}}}$ are roughly proportional to the number of available annotations for training, with the most common concept type ({\sc Mutation}) receiving the smallest boost.

For both models, their performance across the different concept types was roughly proportional to the number of annotations for training.
While there were more {\sc Mutation} annotations than {\sc Cancer} annotations, there was a far greater variety of {\sc Mutation} mentions than {\sc Cancer} mentions, which likely explains why {\sc Cancer} outperforms {\sc Mutation} in both models by roughly 10 points of F1.
{\sc Treatment} is the next most common concept type, and while for BioBERT$_{\textrm{{\sc base}}}$ this performs 6.73 points of F1 worse than {\sc Mutation}, for BioBERT$_{\textrm{{\sc large}}}$ {\sc Treatment} actually outperforms {\sc Mutation} by 1.35 points of F1.
Meanwhile, for both models {\sc Population} is the second-worst-performing concept type, while as mentioned neither model correctly identifies a single {\sc Outcome}.
The latter is almost certainly due to the combination of few annotations (130 in the entire corpus) and long, complex nature of each concept span (28.5 tokens).
Clearly, {\sc Outcome} extraction is not an ideal named entity recognition task and should be handled by a different type of extraction (e.g., sentence classification).

Finally, it is interesting that with the exception of {\sc Population} for BioBERT$_{\textrm{{\sc large}}}$, all concepts have higher recall than precision.
This requires further investigation, but one possibility is that the BERT models are good at identifying instances very similar to those in the training data, but additionally predict spans with high biomedical similarity that are nonetheless not one of the annotated concepts.

\section{Discussion}
\label{section:discussion}

This work is an initial feasibility study on the extraction of key variables for precision oncology from biomedical literature abstracts.
We focus on identifying the type of cancer, mutation, population information, treatment, and outcomes.
A small corpus of 250 abstracts was manually annotated, then two BioBERT models were evaluated.
While none of the five concept types performed up to the level one would hope, {\sc Cancer} performed reasonably well (F1 of 75.00), while {\sc Mutation} and {\sc Treatment} showed promise (F1 of 64.94 and 66.29, respectively).
{\sc Population} performed below a level that is likely usable (F1 of 52.94), while {\sc Outcome} was not successfully extracted at all.
Here, we discuss the successes and shortcomings of this feasibility pilot and what should come next to address the key problems.

The most obvious need for improvement is the small size of the dataset.
Our point of reference for appropriate dataset sizes is the NCBI Disease Corpus,\cite{Dogan12.bionlp,Dogan14.jbi}
which has 793 abstracts, or roughly three times the size of what is presented here.
BioBERT's performance on that corpus is an F1 of 89.71, which we can assume is a rough upper bound for automatic extraction if the corpus was scaled up.
We will note, however, that even the {\sc Cancer}, {\sc Mutation}, and {\sc Treatment} concepts themselves are more diverse than what is in the NCBI Disease Corpus, and the lexical variation seen with even these concepts is likely greater (especially {\sc Treatment}, see Figure~\ref{figure:examples}), so this would be an ambitious upper bound.
Ultimately, it seems clear that increasing the corpus size would be beneficial.

Regarding the lower-performance concepts, it is likely that {\sc Population} needs to be refined as a concept, which would allow it to incorporate pre-defined lexicons.
In this study we intentionally did not define this concept narrowly in order to assess the range of populations mentioned in abstracts.
Going forward, however, we can focus on the set of populations that are critically important to precision oncology.
These usually differ from the normal medical notion of a population.
Instead of demographics, in precision oncology the cancer and treatment history are primary populations of interest (e.g., ``{\em treatment-naive}'' in Figure~\ref{figure:examples} refers to patients who have not yet undergone chemotherapy).
Regarding {\sc Outcome}, this is clearly an item that is more appropriately tackled as a sentence classifier than via entity extraction.
As can be seen in Figure~\ref{figure:examples}, the {\sc Outcome} sentences have fairly clear features not seen in the other sentences, so it is likely that a sentence classifier could identify these with relatively high efficacy.

The comparison of BioBERT$_{\textrm{{\sc base}}}$ and BioBERT$_{\textrm{{\sc large}}}$ is instructive.
At the current size of the corpus, the larger model provides more than sufficient benefit to justify its additional complexity.
Perhaps in a larger corpus, the base model will close the gap.
In other works (e.g., Ji et al.\cite{Ji20.amia_summit}),
the larger model performed no better than the base model.
These experiments, then, should be revisited with a larger corpus.

Another logical place for improvement is the use of knowledge resources.
In this study, we hoped to assess the performance of BioBERT alone, but future work should incorporate existing knowledge resources such as the NCI Thesaurus\cite{Sioutos07} for cancer names and COSMIC\cite{Forbes11} for gene mutations.
Above, we stated the NCBI Disease Corpus performance is a good estimate of an upper bound, but the one advantage of focusing exclusively on precision oncology is that more detailed knowledge resources can be brought to bear: a more specific domain allows us to make domain-specific assumptions.
This could be critical for improving performance, but there is one important note of caution which also justifies our initial reasoning to evaluate a resource-free approach.
Since precision oncology moves quickly as a field, the lexicon of terms used in papers is oftentimes well ahead of knowledge resources.
A new oncogene may be identified months or years before it is incorporated into the appropriate knowledge base.
Over-reliance on these knowledge sources may increase the NLP performance on the annotated corpus while simultaneously reducing the model's ability to recognize the very emerging concepts we are most focused on identifying.
Thus, these knowledge resources cannot be integrated naively, and care should be taken in this process.

A final avenue for improvement focuses on the machine learning aspects.
This includes adjusting the tagging scheme--we used BILOU in this study, but given the variance in concept length (see Table~\ref{table:descriptive_stats}) other tagging schemes may be more appropriate.
Not every concept type need use the same tagging scheme, either. E.g., the shorter {\sc Mutation} concepts may utilize a more simple BIO scheme.
Additionally, the only form of transfer learning we experimented with in this paper is the use of the BioBERT model itself, which effectively transfers a language model pre-trained on large amounts of biomedical text.
After the language modeling, but prior to fine-tuning the model on this precision oncology corpus, other existing datasets may be utilized for transfer learning, such as the NCBI Disease Corpus\cite{Dogan12.bionlp,Dogan14.jbi} and PGxCorpus\cite{Legrand20}.
This would effectively reduce the need to scale up the size of our own manual corpus, though we do not believe that even with transfer learning the current corpus size is sufficient.

\vspace{-0.1in}

\paragraph{Limitations}

The data evaluated in this study was taken from the TREC Precision Medicine track,\cite{Roberts17.trec} and specifically the subset of abstracts marked as relevant for one of the topics.
As such, it is certainly not representative of the full array of biomedical literature.
This decision was made for annotation convenience--these abstracts were known to be highly relevant to precision oncology.
However, the real bias introduced here is the manual nature in which they were chosen.
Identifying potentially relevant abstracts to annotate via keywords or machine learning would result in a corpus that is more appropriate, as these methods could be re-applied when using the precision oncology NLP model on new abstracts.
A second limitation is the training of the annotators was intentionally kept minimal so as to encourage exploration of potential concepts.
Also, only one of the two annotators was biomedically trained.
We have discussed the need for additional manual annotation, but this will also need to come with additional training and more refined guidelines to ensure annotation quality.

\section{Conclusion}
\label{section:conclusion}

This work presents a pilot study for NLP information extraction of terms related to precision oncology from biomedical literature abstracts.
Five concept types were targeted: {\sc Cancer}, {\sc Mutation}, {\sc Population}, {\sc Treatment}, and {\sc Outcome}.
A small corpus of 250 abstracts was manually annotated and reconciled.
Two BioBERT models were evaluated for automatic extraction, with the best results ranging in F1 of 75.0 (for {\sc Cancer}) to a complete inability to extract {\sc Outcome} information.
We finally discussed a set of opportunities for future work to improve these results, including a larger corpus, use of existing biomedical knowledge resources, and additional transfer learning.

\section*{Acknowledgments}
This work was supported by the the Patient-Centered Outcomes Research Institute (PCORI) under award ME‐2018C1‐10963.
The underlying TREC Precision Medicine data was supported by the National Institute of Standards \& \hbox{Technology} (NIST).

\bibliographystyle{vancouver-mod}
\bibliography{all}

\begin{thebibliography}{10}

\bibitem{Collins15}
Collins FS, Varmus H.
\newblock {A New Initiative on Precision Medicine}.
\newblock New England Journal of Medicine. 2015;372:793--795.

\bibitem{Garraway18}
Garraway LA, Verweij J, Ballman KV.
\newblock {Precision Oncology: An Overview}.
\newblock Journal of Clinical Oncology. 2013;31(15).

\bibitem{Hewett02}
Hewett M, Oliver DE, Rubin DL, Easton KL, Stuart JM, Altman RB, Klein TE.
\newblock {PharmGKB: the Pharmacogenetics Knowledge Base}.
\newblock Nucleic Acids Research. 2002;30(1):163--165.

\bibitem{Barbarino18}
Barbarino JM, Whirl‐Carrillo M, Altman RB, Klein TE.
\newblock {PharmGKB: A worldwide resource for pharmacogenomic information}.
\newblock WIREs Systems Biology and Medicine. 2018;10(4):e1417.

\bibitem{Fountzilas18}
Fountzilas E, Tsimberidou AM.
\newblock {Overview of precision oncology trials: challenges and
  opportunities}.
\newblock Expert Review of Clinical Pharmacology. 2018;11(8):797--804.

\bibitem{Chakravarty17}
Chakravarty D, Gao J, Phillips S, Kundra R, Zhang H, Wang J, Rudolph JE, Yaeger
  R, Soumerai T, Nissan MH, Chang MT, Chandarlapaty S, Traina TA, Paik PK, Ho
  AL, et~al.
\newblock {OncoKB: A Precision Oncology Knowledge Base}.
\newblock JCO Precision Oncology. 2017;1.

\bibitem{Roberts17.trec}
Roberts K, Demner-Fushman D, Voorhees EM, Hersh WR, Bedrick S, Lazar A, Pant S.
\newblock {Overview of the TREC 2017 Precision Medicine Track}.
\newblock In: Proceedings of the Twenty-Sixth Text Retrieval Conference; 2017.
  .

\bibitem{Roberts18.trec}
Roberts K, Demner-Fushman D, Voorhees EM, Hersh WR, Bedrick S, Lazar A.
\newblock {Overview of the TREC 2018 Precision Medicine Track}.
\newblock In: Proceedings of the Twenty-Seventh Text Retrieval Conference;
  2018. .

\bibitem{Roberts19.trec}
Roberts K, Demner-Fushman D, Voorhees EM, Hersh WR, Bedrick S, Lazar A.
\newblock {Overview of the TREC 2019 Precision Medicine Track}.
\newblock In: Proceedings of the Twenty-Eighth Text Retrieval Conference; 2019.
  .

\bibitem{Shenoi20.amia_summit}
Shenoi SJ, Ly V, Soni S, Roberts K.
\newblock {Developing a Search Engine for Precision Medicine}.
\newblock In: Proceedings of the AMIA Informatics Summit; 2020. p. 579--588.

\bibitem{Devlin18}
Devlin J, Chang M, Lee K, Toutanova K.
\newblock {BERT: Pre-training of Deep Bidirectional Transformers for Language
  Understanding}.
\newblock arXiv. 2018;abs/1810.04805.
\newblock Available from: \url{http://arxiv.org/abs/1810.04805}.

\bibitem{Lee20}
Lee J, Yoon W, Kim S, Kim D, Kim S, So CH, Kang J.
\newblock {BioBERT: a pre-trained biomedical language representation model for
  biomedical text mining}.
\newblock Bioinformatics. 2020;36(4):1234--1240.

\bibitem{Novichkova03}
Novichkova S, Egorov S, Daraselia N.
\newblock {MedScan, a natural language processing engine for MEDLINE
  abstracts}.
\newblock Bioinformatics. 2003;19(13):1699--1706.

\bibitem{Chun06}
Chun H, Tsuruoka Y, Kim J, Shiba R, Nagata N, Hishiki T, Tsujii J.
\newblock {Automatic recognition of topic-classified relations between prostate
  cancer and genes using MEDLINE abstracts}.
\newblock BMC Bioinformatics. 2006;7:S4.

\bibitem{Baker16.bioinformatics}
Baker S, Silins I, Guo Y, Ali I, H\"{o}gberg J, Stenius U, Korhonen A.
\newblock {Automatic semantic classification of scientific literature according
  to the hallmarks of cancer}.
\newblock Bioinformatics. 2016;32(3):432--440.

\bibitem{Baker16.biotxtm}
Baker S, Korhonen A, Pyysalo S.
\newblock {Cancer Hallmark Text Classification Using Convolutional Neural
  Networks}.
\newblock In: Proceedings of the Fifth Workshop on Building and Evaluating
  Resources for Biomedical Text Mining (BioTxtM 2016); 2016. .

\bibitem{Pyysalo19.bioinformatics}
Pyysalo S, Baker S, Ali I, Haselwimmer S, Shah T, Young A, Guo Y, H\"{o}gberg
  J, Stenius U, Narita M, Korhonen A.
\newblock {LION LBD: a literature-based discovery system for cancer biology}.
\newblock Bioinformatics. 2019;35(9):1553--1561.

\bibitem{Spasic14}
Spasi\'{c} I, Livsey J, Keane JA, Nenadi\'{c} G.
\newblock {Text mining of cancer-related information: review of current status
  and future directions}.
\newblock International Journal of Medical Informatics. 2014;83(9):605--623.

\bibitem{Yim16}
Yim W, Yetisgen M, Harris WP, Kwan SW.
\newblock {Natural Language Processing in Oncology: A Review}.
\newblock JAMA Oncology. 2016;2(6):797--804.

\bibitem{Datta19.jbi}
Datta S, Bernstam EV, Roberts K.
\newblock {A frame semantic overview of NLP-based information extraction for
  cancer-related EHR notes}.
\newblock Journal of Biomedical Informatics;100:103301.

\bibitem{Rindflesch00.biocomputing}
Rindflesch TC, Tanabe L, Weinstein JN, Hunter L.
\newblock {EDGAR: Extraction of Drugs, Genes And Relations from the Biomedical
  Literature}.
\newblock In: Pacific Symposium on Biocomputing; 2000. p. 517--528.

\bibitem{Libbus04}
Libbus B, Kilicoglu H, Rindflesch TC, Mork JG, Aronson AR.
\newblock {Using Natural Language Processing, LocusLink and the Gene Ontology
  to Compare OMIM to MEDLINE}.
\newblock In: HLT-NAACL 2004 Workshop: Linking Biological Literature,
  Ontologies and Databases; 2004. p. 69--76.

\bibitem{GOConsortium04}
Consortium GO.
\newblock {The Gene Ontology (GO) database and informatics resource}.
\newblock Nucleic Acids Research. 2004;32(suppl\_1):D258--D261.

\bibitem{Ahlers07}
Ahlers CB, Fiszman M, Demner-Fushman D, Lang FM, Rindflesch TC.
\newblock {Extracting Semantic Predications from MEDLINE Citations for
  Pharmacogenomics}.
\newblock In: Pacific Symposium on Biocomputing; 2007. .

\bibitem{Klein01}
Klein TE, Chang JT, Cho MK, Easton KL, Fergerson R, Hewett M, Lin Z, Liu Y, Liu
  S, Oliver DE, Rubin DL, Shafa F, Stuart JM, Altman RB.
\newblock {Integrating genotype and phenotype information: an overview of the
  PharmGKB project}.
\newblock The Pharmacogenomics Journal. 2001;1:167--170.

\bibitem{WhirlCarrillo12}
Whirl‐Carrillo M, McDonagh EM, Hebert JM, Gong L, Sangkuhl K, Thorn CF,
  Altman RB, Klein TE.
\newblock {Pharmacogenomics Knowledge for Personalized Medicine}.
\newblock Clinical Pharmacology \& Therapeutics. 2012;92(4):414--417.

\bibitem{Thorn13}
Thorn CF, Klein TE, Altman RB.
\newblock {PharmGKB: The Pharmacogenomics Knowledge Base}.
\newblock In: Innocenti F, {van Schaik} R, editors. Pharmacogenomics. vol.
  1015. Humana Press; 2013. .

\bibitem{Pakhomov12}
Pakhomov S, McInnes BT, Lamba J, Liu Y, Melton GB, Ghodke Y, Lamba NBV,
  Birnbaum AK.
\newblock {Using PharmGKB to train text mining approaches for identifying
  potential gene targets for pharmacogenomic studies}.
\newblock Journal of Biomedical Informatics. 2012;45(5):862--869.

\bibitem{Buyko12}
Buyko E, Beisswanger E, Hahn U.
\newblock {The Extraction of Pharmacogenetic and Pharmacogenomic Relations--A
  Case Study Using PharmGKB}.
\newblock In: Pacific Symposium on Biocomputing; 2012. p. 376--387.

\bibitem{Ravikumar14}
Ravikumar KE, Wagholikar KB, Liu H.
\newblock {Towards Pathway Curation Through Literature Mining-A Case Study
  Using PharmGKB}.
\newblock In: Pacific Symposium on Biocomputing; 2014. p. 352--363.

\bibitem{Legrand20}
Legrand J, Gogdemir R, Bousquet C, Dalleau K, Devignes M, Digan W, Lee C,
  Ndiaye N, Petitpain N, Ringot P, {Sma\"{i}l-Tabbone} M, Toussaint Y, Coulet
  A.
\newblock {PGxCorpus, a manually annotated corpus for pharmacogenomics}.
\newblock Scientific Data. 2020;7:3.

\bibitem{Cohen17.hla}
Cohen KB, Verspoor K, Fort K, Funk C, Bada M, Palmer M, Hunter LE.
\newblock {The Colorado Richly Annotated Full Text (CRAFT) Corpus: Multi-Model
  Annotation in the Biomedical Domain}.
\newblock In: Ide N, Pustejovsky J, editors. Handbook of Linguistic Annotation;
  2017. p. 1379--1394.

\bibitem{Baumgartner19}
Baumgartner W, Bada M, Pyysalo S, Ciosici MR, Hailu N, {Pielke-Lombardo} H,
  Regan M, Hunter L.
\newblock {CRAFT Shared Tasks 2019 Overview {---} Integrated Structure,
  Semantics, and Coreference}.
\newblock In: Proceedings of The 5th Workshop on BioNLP Open Shared Tasks;
  2019. p. 174--184.

\bibitem{Deng19.jcocci}
Deng Z, Yin K, Bao Y, Armengol VD, Wang C, Tiwari A, Barzilay R, Parmigiani G,
  Braun D, Hughes KS.
\newblock {Validation of a Semiautomated Natural Language Processing–Based
  Procedure for Meta-Analysis of Cancer Susceptibility Gene Penetrance}.
\newblock JCO Clinical Cancer Informatics. 2019;3.

\bibitem{Bao19.jcocci}
Bao Y, Deng Z, Wang Y, Kim H, Armengo VD, Acevedo F, Ouardaoui N, Wang C,
  Parmigiani G, Barzilay R, Braun D, Hughes KS.
\newblock {Using Machine Learning and Natural Language Processing to Review and
  Classify the Medical Literature on Cancer Susceptibility Genes}.
\newblock JCO Clinical Cancer Informatics. 2019;3.

\bibitem{Hughes20.breast}
Hughes KS, Zhou J, Bao Y, Singh P, Wang J, Yin K.
\newblock {Natural language processing to facilitate breast cancer research and
  management}.
\newblock The Breast Journal. 2020;26:92--99.

\bibitem{Stenetorp12}
Stenetorp P, Pyysalo S, Topi\'{c} G, Ohta T, Ananiadou S, Tsujii J.
\newblock {brat: a web-based tool for NLP-assisted text annotation}.
\newblock In: Proceedings of the Demonstration Session at EACL 2012; 2012. p.
  102--107.

\bibitem{spacy2}
Honnibal M, Montani I.
\newblock {spaCy 2}: Natural language understanding with {B}loom embeddings,
  convolutional neural networks and incremental parsing; 2017.

\bibitem{Dogan12.bionlp}
Do\u{g}an RI, Lu Z.
\newblock {An improved corpus of disease mentions in PubMed citations}.
\newblock In: Proceedings of the 2012 Workshop on Biomedical Natural Language
  Processing;. p. 91--99.

\bibitem{Dogan14.jbi}
Do\u{g}an RI, Leaman R, Lu Z.
\newblock {NCBI disease corpus: a resource for disease name recognition and
  concept normalization}.
\newblock Journal of Biomedical Informatics. 2014;47:1--10.

\bibitem{Ji20.amia_summit}
Ji Z, Wei Q, Xu H.
\newblock {BERT-based Ranking for Biomedical Entity Normalization}.
\newblock In: Proceedings of the AMIA Joint Summits on Translational Science;
  2020. p. 269--277.

\bibitem{Sioutos07}
Sioutos N, {de Coronado} S, Haber MW, Hartel FW, Shaiu W, Wright LW.
\newblock {NCI Thesaurus: A semantic model integrating cancer-related clinical
  and molecular information}.
\newblock Journal of Biomedical Informatics. 2007;40(1):30--43.

\bibitem{Forbes11}
Forbes SA, Bindal N, Bamford S, Cole C, Kok CY, Beare D, Jia M, Shepherd R,
  Leung K, Menzies A, Teague JW, Campbell PJ, Stratton MR, Futreal PA.
\newblock {COSMIC: mining complete cancer genomes in the Catalogue of Somatic
  Mutations in Cancer}.
\newblock Nucleic Acids Research. 2011;39(Database issue):D945--D950.

\end{thebibliography}

\end{document}